\documentclass[10pt, a4paper]{article}
\usepackage{lrec2022} 
\usepackage{multibib}
\newcites{languageresource}{Language Resources}
\usepackage{graphicx}
\usepackage{tabularx}
\usepackage{soul}
\usepackage{titlesec}
\titleformat{\section}{\normalfont\large\bfseries\center}{\thesection.}{1em}{}
\titleformat{\subsection}{\normalfont\SmallTitleFont\bfseries\raggedright}{\thesubsection.}{1em}{}
\titleformat{\subsubsection}{\normalfont\normalsize\bfseries\raggedright}{\thesubsubsection.}{1em}{}
\renewcommand\thesection{\arabic{section}}
\renewcommand\thesubsection{\thesection.\arabic{subsection}}
\renewcommand\thesubsubsection{\thesubsection.\arabic{subsubsection}}

\usepackage{epstopdf}
\usepackage[utf8]{inputenc}
\usepackage{hyperref}
\usepackage{xstring}
\usepackage{color}
\usepackage[T1]{fontenc}
\usepackage{microtype}

\title{Predicting Embedding Reliability in Low-Resource Settings \\ Using Corpus Similarity Measures}

\name{Jonathan Dunn, Haipeng Li, Damian Sastre} 

\address{University of Canterbury, Department of Linguistics \\ and the New Zealand Institute for Language, Brain and Behaviour \\
         Christchurch, New Zealand \\
         \href{mailto:jonathan.dunn@canterbury.ac.nz}{jonathan.dunn@canterbury.ac.nz}, \href{mailto:haipeng.li@canterbury.ac.nz}{haipeng.li@canterbury.ac.nz}, \href{mailto:dsa105@uclive.ac.nz}{dsa105@uclive.ac.nz} \\
         }

\abstract{
This paper simulates a low-resource setting across 17 languages in order to evaluate embedding similarity, stability, and reliability under different conditions. The goal is to use corpus similarity measures before training to predict properties of embeddings after training. The main contribution of the paper is to show that it is possible to predict downstream embedding similarity using upstream corpus similarity measures. This finding is then applied to low-resource settings by modelling the reliability of embeddings created from very limited training data. Results show that it is possible to estimate the reliability of low-resource embeddings using corpus similarity measures that remain robust on small amounts of data. These findings have significant implications for the evaluation of truly low-resource languages in which such systematic downstream validation methods are not possible because of data limitations. \\ \newline  }

\begin{document}

\maketitleabstract

\section{Validating Low-Resource Embeddings}

This paper simulates a low-resource setting for 17 non-English languages in order to evaluate embedding similarity, stability, and reliability under different conditions. While these are actually high-resource languages, we are able to simulate a low-resource setting across different language families, writing systems, and types of morphology by constraining both the amount and the type of data that is made available. We then try to predict the downstream similarity, stability, and reliability of embeddings given upstream properties of the training corpora. The larger goal is to predict the reliability of embeddings in truly low-resource settings in which such experiments are not possible. The key finding of the paper is that there is a strong relationship between the similarity of training corpora and the similarity of embeddings, a relationship that extends across a diverse range of languages and registers.

Embeddings remain a key representation within \textsc{nlp}, as shown by many samples of recent work \cite{miaschi-dellorletta-2020-contextual,adelmann-etal-2021-impact}. At the same time, however, recent work has also shown that embeddings are surprisingly variable \cite{wendlandt-etal-2018-factors,burdick-etal-2021-analyzing}. Such work has shown that high-resource languages like English, with many billions of words available for training, have embeddings that differ by data set \cite{antoniak-mimno-2018-evaluating}, by geographic population \cite{dunn-adams-2020-geographically}, and even by random iterations on the same data set \cite{Hellrich2019}. The basic implication is that, for instance, English embeddings from web data from South Asia are expected to be quite different from English embeddings from American news articles. 

For high-resource languages, such variability is mitigated by the wide availability of in-domain training data for most tasks. But for low-resource languages there is a systematic gap in the kind of training data that is available. For instance, many languages have the Bible \cite{ChristodoulopoulosC.andSteedman2015} or related religious literature \cite{agic-vulic-2019-jw300} as their largest corpus. The problem is that representations learned from such corpora are likely to be significantly different from those learned from other sources.

How much do representations of low-resource languages depend on the selection of data that happens to be available? To answer this question, this paper uses corpus similarity measures on training data (upstream) to predict differences in trained embeddings (downstream). The basic idea is to model the influence of training data on the variability of embeddings by simulating different low-resource contexts. 

This question is important because most languages are relatively low-resource, lacking data sets that contain billions of words. The ability to predict variability in embeddings given training data would enable us to estimate reliability in low-resource languages for which evaluations such as those in this paper are not possible.

\section{Experimental Questions}

\begin{table*}[t]
\centering
\begin{tabular}{|p{2.8cm}|p{1cm}|p{3cm}|p{3cm}|p{3cm}|}
\hline
\textbf{Language} & \textbf{Code} & \textbf{Family} & \textbf{Writing} & \textbf{Morphology} \\
\hline
Arabic & ara & Afro-Asiatic & Abjad & Root-Pattern \\
Indonesian & ind & Austronesian & Alphabet & Agglutinative \\
Polish & pol & IE:Balto-Slavic & Alphabet & Fusional \\
Russian & rus & IE:Balto-Slavic & Alphabet & Fusional \\
German & deu & IE:Germanic & Alphabet & Fusional \\
Dutch & nld & IE:Germanic & Alphabet & Analytic \\
Swedish & swe & IE:Germanic & Alphabet & Analytic \\
Greek & ell & IE:Hellenic & Alphabet & Fusional \\
Farsi & fas & IE:Indo-Iranian & Abjad & Analytic \\
French & fra & IE:Romance & Alphabet & Fusional \\
Italian & ita & IE:Romance & Alphabet & Fusional \\
Portuguese & por & IE:Romance & Alphabet & Fusional \\
Spanish & spa & IE:Romance & Alphabet & Fusional \\
Japanese & jpn & Isolate & Logographic & Agglutinative \\
Korean & kor & Isolate & Logographic & Agglutinative \\
Turkish & tur & Turkic & Alphabet & Agglutinative \\
Finnish & fin & Uralic & Alphabet & Agglutinative \\
\hline
  \end{tabular}
  \caption{Languages Used in Experiments, Sorted By Family, with Writing System and Type of Morphology}
  \label{tab:1}
\end{table*}

The main contribution of this paper is to evaluate the influence of training corpora on embedding stability for low-resource languages by simulating low-resource and medium-resource settings. We use measures of corpus similarity to determine both (i) relationships between sets of training data and (ii) homogeneity within individual training sets. The basic question is whether we can predict downstream embedding similarity (after training) given upstream corpus similarity (before training). The larger goal is to estimate the reliability of embeddings for low-resource languages, in which systematic evaluations of different downstream embeddings is not possible because of insufficient data. This first section introduces the main experimental conditions and the questions they are used to address.

\textbf{Source}. How does the source of training data impact embedding similarity? We draw training data from three distinct registers: social media, Wikipedia, and web pages. A register is a unique context of production associated with a specific communicative situation \cite{Biber2009}. A long line of research has shown that register has a significant impact on both grammar and the lexicon \cite{Biber2012,Biber2020}. Because of the significance of register variation, we expect that embeddings trained from different registers (such as tweets vs Wikipedia articles) will themselves be quite different. While high-resource languages have many registers available for training purposes, low-resource languages often have data from a limited range of registers (e.g., religious or legal documents). This experimental condition, register-specific embeddings, allows us to evaluate whether the context of production has a significant influence on downstream embeddings.

\textbf{Size}. How does the amount of training data impact embedding variability? We evaluate embedding stability over increasing amounts of training data in order to determine whether more data overall is able to compensate for differences in the data. This condition looks at corpora ranging from 10 million words to 100 million words in increments of 10 million. This line of experimentation allows us to simulate low-resource and medium-resource contexts to find out how embeddings change given more training data.

\textbf{Language Properties}. Do specific types of languages have more stable embeddings? The experiments here are conducted across 17 non-English languages as shown in Table \ref{tab:1}. These languages represent 10 unique sub-families, three types of writing system, and four types of morphology. This selection of languages allows us to determine if any of the observed behaviours can be attributed to a specific type of language.

In the next section, we position this current study against related work. We then present the underlying data sets used in the experiments (Section 4) and the methods used for training embeddings and calculating both corpus similarity and embedding similarity (Section 5). We then analyze the impact of different registers (Section 6), the impact of increasing amounts of training data within registers (Section 7), the reliability of low-resource embeddings (Section 8), and the reliability of corpus similarity measures (Section 9). Finally, we consider the implications of this work for natural language processing more broadly (Section 10).

\section{Related Work}

\begin{figure*}[t]
\centering
\includegraphics[width = 450pt]{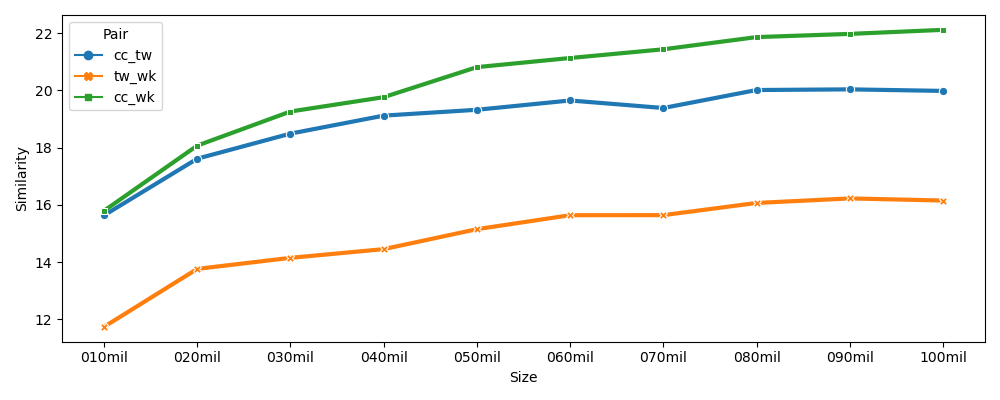}
\caption{Embedding Similarity Across Registers by Increased Training Data, Arabic}
\label{fig1}
\end{figure*}

This section reviews related work on both corpus similarity and embedding stability. First, corpus similarity measures have been used in different applications, such as text classification, information retrieval, and the evaluation of machine translation. Originally defined as a problem within corpus linguistics \cite{kilgarriff-1997-using,Kilgarriff2001}, many measures have since been proposed \cite{kilgarriff-2009,fothergill-etal-2016-evaluating,Piperski-2017,lu-etal-2020-diverging}. 

One common feature across these approaches is that they are based upon word frequency; in fact, frequency-based approaches have consistently out-performed model-based approaches \cite{fothergill-etal-2016-evaluating}. More recently, these measures have been used to evaluate fluctuation within and between registers for different language varieties \cite{Dunn2021}, as well as to classify documents \cite{Nanayakkara-2018,DBLP:conf/ecir/LebanFG16} and detect paraphrases in German \cite{DBLP:journals/corr/Torres-MorenoSP17}. 

In terms of embedding stability, recent work has used word similarity to investigate variation between embeddings trained on a single corpus \cite{antoniak-mimno-2018-evaluating}, focusing on the training corpus itself as a source of variation. After examining four algorithms and six data sets, results show that corpus size is one of several sources of variability between embeddings. The broader conclusion, shared by the experiments in this paper, is that embeddings represent a specific corpus rather than representing an entire language. 

Other work has focused on evaluating whether various factors contribute to the stability of word embeddings and analysing the effects of stability on downstream tasks \cite{wendlandt-etal-2018-factors}. Using a ridge regression model to predict the stability of individual words, these results show that stability within domains is greater than stability across domains. \textit{Domains} in this setting are comparable to registers. 

To evaluate the stability of word embeddings derived from a single corpus, \cite{Hellrich2019} modifies the Singular Value Decomposition algorithm and compares it with other algorithms on three English corpora that ultimately represent distinct registers. The results show that the modified SVD is found to be both reliable and accurate as compared to other algorithms. This study also concludes that stability is positively influenced by corpus size, so that larger sizes lead to higher stability.

Recent work has used linguistic properties to explain the stability of word embeddings across different languages \cite{burdick-etal-2021-analyzing}. Again using a regression model, this work finds that languages with more complex morphology tend to be less stable than languages with simpler morphology. That finding is not replicated in this present study, although the issue is raised by the case of Arabic in Section 8. Most other work, however, focuses only on English \cite{antoniak-mimno-2018-evaluating,wendlandt-etal-2018-factors,Hellrich2019}. 

This paper is unique in its cross-linguistic, cross-register, and cross-size experimental design. This approach allows us to determine in a systematic manner which properties of corpora influence embedding stability.

\begin{figure*}[t]
\centering
\includegraphics[width = 450pt]{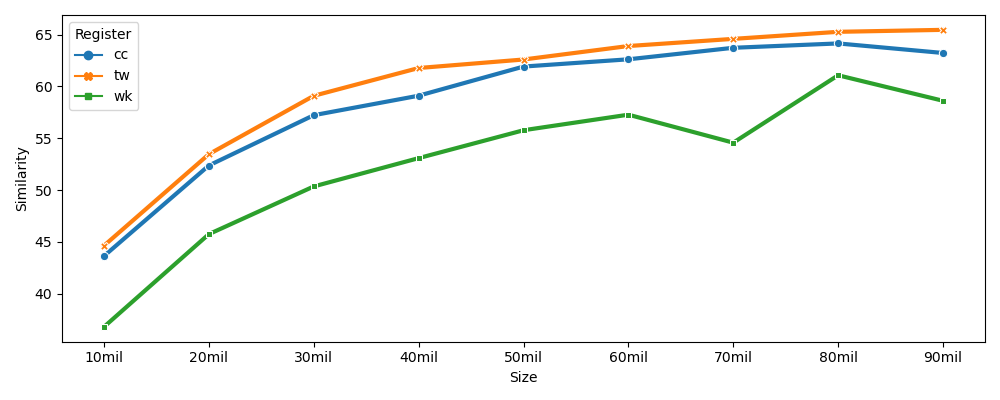}
\caption{Embedding Stability Within Registers by Increased Training Data, Arabic}
\label{fig2}
\end{figure*}

\section{Data}

The data for these experiments comes from comparable corpora representing 17 languages. The Wikipedia register (\textsc{wk}) is collected from the public Wikimedia dump of March 2020. Languages are identified using the designation provided by Wikipedia. The web register (\textsc{cc}) is collected from the Corpus of Global Language Use \cite{Dunn2020}, ultimately derived from the Common Crawl. The social media register (\textsc{tw}) is collected from geo-referenced tweets. For both web pages and tweets, languages are identified using the idNet package \cite{Dunn2020}. A list of languages is provided in Table \ref{tab:1}, including the family, type of writing system, and type of morphology for each language. Each source of data (i.e., register) provides a corpus of 100 million words.

\section{Methods}

The experiments in this paper depend on two measures: similarity between embeddings and similarity between training corpora. Following previous work \cite{wendlandt-etal-2018-factors}, we calculate the similarity between two sets of embeddings by taking the aggregate overlap of nearest neighbors. This is calculated as follows: first, we create an independent corpus for each language, representing different registers from the main experimental data. These background corpora contain movie subtitles, news commentary articles, and Bible translations \cite{Tiedemann2012,ChristodoulopoulosC.andSteedman2015}. Thus, each language is represented by the three registers described above (\textsc{wk, tw, cc}) in addition to these separate out-of-domain corpora. We find the 1,000 most common lexical items in these background corpora. For each of the top lexical items, we then retrieve the ten nearest neighbors from each set of embeddings. 

The overlap for each lexical item is the percentage of words which appear as neighbors within both sets of embeddings. For example, if \textit{dog} is a nearest neighbor for \textit{cat} in both tweet-based and web-based embeddings, this indicates a certain similarity between those sets of embeddings. Thus, higher overlap scores indicate higher agreement. The overall embedding similarity between two conditions (such as tweets vs Wikipedia at 50 million words) is represented using the average overlap across these top 1,000 words.

The embeddings themselves are character-based, trained using the skip-gram negative sampling method with 50 negative examples per observation and trained for 20 epochs. Implemented using the fastText framework \cite{mikolov2018advances}, these popular embeddings have the advantage of being character-based, which we expect to reduce differences that are caused by morphology or writing system (c.f., Table 1).

We also need to measure the similarity between the training corpora themselves, before training takes place \cite{kilgarriff-1997-using,fothergill-etal-2016-evaluating}. Recent work on corpus similarity measures has shown that a frequency-based approach with 5k bag-of-words features and Spearman’s \textit{rho} performs well across many languages \cite{Dunn2021}. A frequency-based approach to corpus similarity uses a vector of n-gram frequencies, usually restricted to the most common n-grams. Spearman's \textit{rho} has been shown to be highly accurate in comparing these frequency vectors, with more similar corpora having a higher correlation coefficient. Unlike other measures \cite{Kilgarriff2001}, Spearman's \textit{rho} is not dependent on corpus size. We use the same background corpora as above to select the top n-gram features for each language.

\begin{table}[t]
\centering
\begin{tabular}{|l|l|c|c|}
\hline
\textbf{Language} & \textbf{Family} & \textbf{Features} & \textbf{Acc.} \\
\hline
Arabic & Afro-Asiatic & C4 & 99\% \\
German & IE:Germanic & C4 & 98\% \\
Greek & IE:Hellenic & W1 & 97\% \\
Farsi & IE:Indo-Iranian & W1 & 96\% \\
Finnish & Uralic & C4 & 94\% \\
French & IE:Romance & W1 & 100\% \\
Indonesian & Austronesian & C4 & 99\% \\
Italian & IE:Romance & W1 & 94\% \\
Japanese & Isolate & C2 & 88\% \\
Korean & Isolate & C4 & 99\% \\
Dutch & IE:Germanic & W1 & 100\% \\
Polish & IE:Balto-Slavic & W1 & 99\% \\
Portuguese & IE:Romance & C4 & 98\% \\
Russian & IE:Balto-Slavic & C4 & 100\% \\
Spanish & IE:Romance & C4 & 99\% \\
Swedish & IE:Germanic & C4 & 96\% \\
Turkish & Turkic & C4 & 100\% \\
\hline
  \end{tabular}
  \caption{Accuracy of Corpus Similarity Measures with Feature Type in a Register Identification Task}
  \label{tab:accuracy}
\end{table}

While the embedding similarity measure has been evaluated before \cite{wendlandt-etal-2018-factors}, we provide an evaluation of the corpus similarity measure here. The underlying task for validation is to predict whether two samples come from the same register: for example, given two samples, do both represent tweets? A high accuracy means that the corpus similarity measures always distinguish between data sources. To convert this continuous measure into a categorical prediction, we set a threshold; the more often this threshold leads to correct predictions, the more accurate the measure is. The threshold calculation takes the lowest average similarity for same-register pairs (for example, \textsc{cc-cc}) and the highest average similarity for cross-register pairs (for example, \textsc{cc-tw}). The threshold is set halfway between these minimum and maximum values \cite{Ali2011,DBLP:conf/ecir/LebanFG16}. We conduct the evaluation using a five-fold cross-validation design, with the threshold calculated on the training data for each fold. Each \textit{sample} in this design is a 20k word sub-set of a corpus.

The accuracy of this register-prediction task is shown in Table \ref{tab:accuracy}, along with the best type of feature (word or character n-gram size). While there is some variation in performance, with Japanese being particularly low, the generally high performance provides a validation of the corpus similarity measure. This is important because it means that these measures are indeed able to distinguish between corpora representing the three registers used in these experiments (\textsc{cc, tw, wk}). In other words, these results show that it is possible to measure differences between training corpora before we train embeddings.

While the corpus similarity measure performs well, the scores are not directly comparable across languages because each language has a different central tendency. For comparison across different sources (i.e., the web vs tweets), we retain the raw similarity value and restrict ourselves to within-language comparisons. For comparison within sources (i.e., the web vs the web), we use the z-score to standardize the measure across all registers, in order to make better downstream predictions (c.f., Section 8). Finally, the similarity between large corpora are estimated by sampling 200 unique pairs of sub-corpora, each containing 20k words. We then find the mean similarity across all samples to represent the relationship between the larger corpora. A Python package for reproducing these corpus similarity measures is available \href{https://github.com/jonathandunn/corpus_similarity}{\underline{here}}.

\section{Experiment 1: Register}

The first experiment asks whether more similar corpora produce more similar embeddings. In other words, is there a relationship between the input (a corpus) and the output (word embeddings)? A different way of asking this same question is whether register variation, a long-studied linguistic phenomenon, has a predictable impact on embeddings. We create three cross-register comparisons for each language: \textsc{cc-tw}, \textsc{tw-wk}, and \textsc{cc-wk}. For each comparison, we compute both the corpus similarity (upstream) and the embedding similarity (downstream). We can visualize embedding similarity in this context as in Figure \ref{fig1}, which shows similarity for each pair of register-specific embeddings (y-axis) over increasing amounts of training data (x-axis) for Arabic. Each line here represents the similarity between two different sets of embeddings. We see, for instance, that all register-specific embeddings become more similar as the amount of training data increases. At the same time, the tweet-based and Wikipedia-based embeddings (in yellow) are much less similar than the others. This indicates that register variation does, in fact, have an impact on these sets of embeddings.

The main question here concerns the similarity between each pair of register-specific embeddings at different data sizes (i.e., at 100 million words). We see across languages that there is a clear set of relationships between embeddings trained on different corpora: register has a significant impact on embeddings, as we expect from previous work on register variation. The full set of figures for each language is provided in the supplementary material, available \href{https://www.dropbox.com/s/hsdzyatb4d2a3bm/LREC_2022.Supplementary_Material.zip?dl=0}{\underline{here}}. The question, however, is whether we can predict these relationships between embeddings using corpus similarity measures.

\begin{table}[t]
\centering
\begin{tabular}{|c|c|c|}
\hline
\textbf{Register Pair} & \textbf{Embedding Sim.} & \textbf{Corpus Sim.} \\
\hline
\textsc{cc-tw} & 24.8 & 0.72 \\
\textsc{tw-wk} & 14.3 & 0.59 \\
\textsc{cc-wk} & 19.6 & 0.66 \\
\hline
  \end{tabular}
  \caption{Relationship between Embedding Similarity and Corpus Similarity for Finnish}
  \label{tab:2}
\end{table}

We take a closer look in Table \ref{tab:2} with Finnish. For each pair of registers, in the first column, we see both the similarity between embeddings and the similarity between the training corpora. For both measures, higher values indicate higher similarity; the scales, however, are quite different. What we see here is that the same pair of registers (\textsc{cc-tw}) produce the most similar embeddings (24.80\% overlap) and also have the highest corpus similarity score (0.72). In fact, there is a strong correlation of 0.999 between these two sets of values. This means that, for Finnish, we can predict which embeddings will be more similar even before we train them.

\begin{table}[t]
\centering
\begin{tabular}{|l|l|c|c|}
\hline
\textbf{Language} & \textbf{Family} & \textbf{10 mil} & \textbf{100 mil} \\
\hline
Arabic & Afro-Asiatic & 0.727 & 0.909 \\
German & IE:Germanic & 0.977 & 0.993 \\
Greek & IE:Hellenic & 0.817 & 0.914 \\
Farsi & IE:Indo-Iranian & 0.591 & 0.729 \\
Finnish & Uralic & 0.988 & 1.000 \\
French & IE:Romance & 0.947 & 0.924 \\
Indonesian & Austronesian & 0.988 & 0.981 \\
Italian & IE:Romance & 0.986 & 0.928 \\
Japanese & Isolate & 0.994 & 0.971 \\
Korean & Isolate & 0.541 & 0.972 \\
Dutch & IE:Germanic & 1.000 & 0.981 \\
Polish & IE:Balto-Slavic & 0.947 & 0.898 \\
Portuguese & IE:Romance & 0.788 & 0.863 \\
Russian & IE:Balto-Slavic & 0.836 & 1.000 \\
Spanish & IE:Romance & 0.838 & 0.966 \\
Swedish & IE:Germanic & 0.999 & 0.995 \\
Turkish & Turkic & 0.678 & 0.906 \\
\hline
\textbf{Average} & \textbf{All} & \textbf{0.861} & \textbf{0.937} \\
\hline
  \end{tabular}
  \caption{Relationship between Embedding Similarity and Corpus Similarity}
  \label{tab:3}
\end{table}

We take a cross-linguistic view of this relationship between input and output in Table \ref{tab:3}. The quantity we are interested in is the relationship between the two measures, corpus similarity and embedding similarity: how well could we predict embedding similarity downstream given corpus similarity upstream? This table shows the relationship both at 10 million words and at 100 million words. Note that the corpus similarity measure remains quite stable across corpus size (c.f., Section 9) while embeddings become more similar given more training data (c.f., Figure \ref{fig1}). We see that the relationship becomes stronger as the embeddings have more training data, from an average correlation of 0.86 to 0.93. This is because the embeddings themselves become more stable with increased training data (c.f., Section 7).

Analyzing Table \ref{tab:3}, we can use the properties of each language to understand what causes specific outliers. For example, the relationship for Korean is quite low at 10 million words but rather high at 100 million words. We might think this is caused by the logographic writing system, but that same pattern is not shown in Japanese. The lowest relationship is shown by Farsi, with a maximum correlation of 0.729. However, we know that this is not caused by the writing system (shared with Arabic) or by the type of morphology (shared by several other languages). This would rather seem to be a specific property of Farsi corpora, rather than a property based on either Farsi's writing system or morphology.

This section has shown two important properties of the impact of register variation on embeddings: First, across all languages there is a significant difference between register-specific embeddings. This means that it is more accurate to formulate Arabic-Wikipedia embeddings rather than universal Arabic embeddings: the downstream embeddings remain register-specific, at least with this amount of training data. In other words, register variation has a consistent impact downstream on trained embeddings. Second, there is a strong relationship between the training corpora themselves and the similarity between embeddings trained on those corpora. In other words, more similar corpora produce more similar sets of embeddings. This is an important finding because it suggests that we should be able to predict the conditions under which embeddings will be both stable and reliable.

\section{Experiment 2: Size}

The second experiment quantifies the amount of change that occurs within register-specific embeddings as the amount of training data is increased. The central question is whether it is possible to predict the growth curve of embedding stability, the rate at which embeddings become more similar when trained from different subsets of the same corpus. For example, Figure \ref{fig2} shows embedding similarity by size within each register for Arabic. Here each line represents a single register, with the comparison made between embeddings trained using different amounts of data. For example, the green line represents Wikipedia. With less data, the agreement between the 10 million and 20 million word conditions (on the left) is below 40\%; but with more data the agreement between the 90 million and 100 million word conditions (on the right) is closer to 60\%. 

We refer to this as \textit{stability} because the two corpora overlap to a large degree. This gives us a baseline for stability within each language: cross-register similarity, for example, should never exceed within-register similarity in this setting. As before, we expect that more training data leads to more stable embeddings; this trend is found across all languages. The more meaningful question, however, is whether we can predict this increasing rate of stability using corpus similarity.

\begin{table}[h]
\centering
\begin{tabular}{|l|c|c|c|c|}
\hline
\textbf{Language} & \textbf{Feature} & \textbf{\textsc{cc}} & \textbf{\textsc{tw}} & \textbf{\textsc{wk}} \\
\hline
Arabic & C4 & 63.22 & 65.45 & 58.61 \\
German & C4 & 64.28 & 67.88 & 59.71 \\
Greek & W1 & 62.99 & 68.21 & 57.59 \\
Farsi & W1 & 62.53 & 68.87 & 57.51 \\
Finnish & C4 & 63.34 & 67.94 & 57.92 \\
French & W1 & 58.99 & 65.63 & 51.71 \\
Indonesian & C4 & 67.77 & 57.90 & 60.79 \\
Italian & W1 & 65.16 & 62.66 & 62.65 \\
Japanese & C2 & 54.81 & 59.42 & 50.29 \\
Korean & C4 & 59.26 & 62.74 & 55.61 \\
Dutch & W1 & 64.56 & 67.17 & 60.43 \\
Polish & W1 & 66.50 & 69.78 & 52.93 \\
Portuguese & C4 & 64.80 & 60.10 & 57.58 \\
Russian & C4 & 57.31 & 67.96 & 60.18 \\
Spanish & C4 & 67.39 & 70.59 & 60.43 \\
Swedish & C4 & 65.02 & 65.64 & 52.89 \\
Turkish & C4 & 61.40 & 66.14 & 54.62 \\
\hline
\textbf{Average} & \textbf{~} & \textbf{62.90} & \textbf{65.53} & \textbf{57.14} \\
\hline
  \end{tabular}
  \caption{Embedding Stability by Language and Register, 90 million and 100 million word comparison}
  \label{size_results}
\end{table}

In other words, we might expect that more homogeneous corpora, data sets that are more self-similar, will produce more stable embeddings because they contain less internal variation. We test this hypothesis in two ways: First, we take the amount of increase in embedding stability for each condition. For example, Arabic web corpora have an increase of 20.49\% in embedding stability between the 10-20 million and 90-100 million word comparisons. But the Arabic Wikipedia corpora have a higher increase of 24.26\%. To test whether we can predict the amount of increased stability, we look at the correlation between (i) the increased stability of within-register embeddings and (ii) the homogeneity of the training corpus. \textit{Homogeneity} here is the same corpus similarity measure calculated across 200 unique chunks from a single corpus. However, there is no consistent relationship. A second approach looks at the relationship between the slope of increased stability across all conditions and corpus homogeneity. Again, there is no consistent relationship across languages.

This experiment therefore reaches a negative result: it is not possible to use corpus homogeneity to predict embedding stability in this context. Table \ref{size_results} shows within-register similarity at the 90-100 million word comparison condition. On the one hand, for each condition there is higher similarity in this same-register comparison than we saw in the previous cross-register comparison, as we would expect. However, it is not possible to predict the rate or the degree of increased embedding stability given corpus homogeneity.

\begin{figure*}[t]
\centering
\includegraphics[width = 450pt]{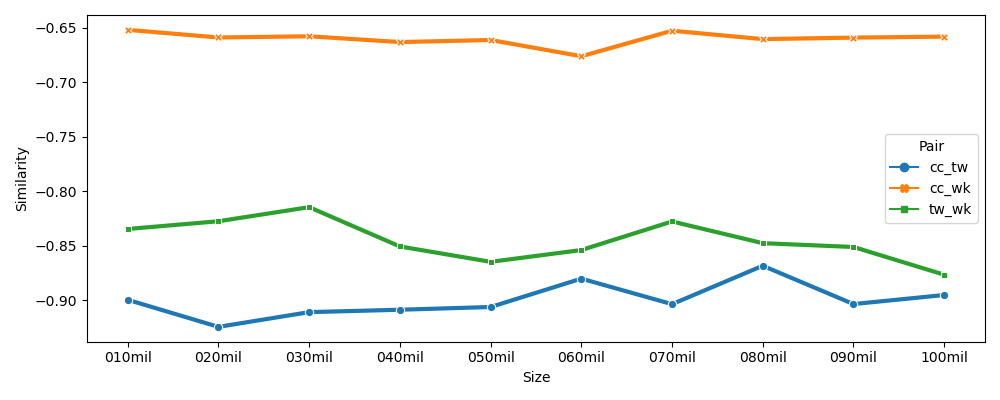}
\caption{Corpus Similarity Between Registers by Increased Training Data, Arabic}
\label{fig3}
\end{figure*}

\section{Experiment 3: Reliability}

The third experiment asks whether we can predict the reliability of embeddings in very low-resource settings. First, we train embeddings using only 1 million words from each language in each register: 1 million words of Arabic tweets, for example. We repeat this process multiple times and calculate similarity between ten unique pairs of same-register embeddings in each language. For many low-resource languages we have only 1 million words available, so that embeddings must be trained on whatever data there is. This experiment simulates many different samples from high-resource languages in order to capture the distribution of embedding similarity values in a low-resource setting: how different would register-specific embeddings have been if we had instead observed some other sub-set of the data?

\begin{table}[t]
\centering
\begin{tabular}{|c|c|c|}
\hline
\textbf{Register} & \textbf{Reliability} & \textbf{Homogeneity} \\
\hline
\textsc{cc} & 21.63 & -0.28 \\
\textsc{tw} & 26.42 & -0.19 \\
\textsc{wk} & 16.88 & -0.36 \\
\hline
  \end{tabular}
  \caption{Relationship between Embedding Reliability and Corpus Homogeneity for Italian}
  \label{tab:4}
\end{table}

As we have seen, larger training sets lead to more stable embeddings. This means, for example, that representations based on a very small amount of data are expected to be less stable. Our question here is about degrees of instability: can we predict which low-resource embeddings will be more reliable across different samples? The basic idea is that, if such predictions are possible, we can estimate reliability in truly low-resource settings for which multiple training sets are not available.

We use Italian as an example in Table \ref{tab:4}, showing the reliability of embeddings in a low-resource setting together with the homogeneity of the training corpora. \textit{Reliability} here is the average agreement across 10 random pairs of embeddings, each representing a unique sub-set of the same corpus. The overall relationship in this case is quite strong, a correlation of 0.999. For example, the most homogeneous corpus (\textsc{tw}), has the highest reliability across random pairs of embeddings. Note that the homogeneity scores have been standardized using the z-score across all register combinations; thus, the mean is 0 with values above 0 indicating high similarity and values below 0 indicating lower similarity. The central question is whether corpus homogeneity can be used to predict the reliability of low-resource embeddings.

\begin{table}[t]
\centering
\begin{tabular}{|l|l|c|}
\hline
\textbf{Language} & \textbf{Family} & \textbf{Correlation} \\
\hline
Arabic & Afro-Asiatic & 0.567 \\
German & IE:Germanic & 0.897 \\
Greek & IE:Hellenic & 0.827 \\
Farsi & IE:Indo-Iranian & 0.993 \\
Finnish & Uralic & 0.847 \\
French & IE:Romance & 0.979 \\
Indonesian & Austronesian & 0.976 \\
Italian & IE:Romance & 0.999 \\
Japanese & Isolate & 0.999 \\
Korean & Isolate & 1.000 \\
Dutch & IE:Germanic & 0.976 \\
Polish & IE:Balto-Slavic & 0.890 \\
Portuguese & IE:Romance & 0.910 \\
Russian & IE:Balto-Slavic & 0.786 \\
Spanish & IE:Romance & 0.999 \\
Swedish & IE:Germanic & 0.962 \\
Turkish & Turkic & 0.972 \\
\hline
\textbf{Average} & \textbf{All} & \textbf{0.916} \\
\hline
  \end{tabular}
  \caption{Relationship between Embedding Reliability and Corpus Homogeneity}
  \label{tab:5}
\end{table}

The relationship between homogeneity and embedding reliability in low-resource settings is shown in Table \ref{tab:5}. This relationship is rather strong on average, 0.916, allowing us to predict the conditions under which embeddings in a low-resource setting will be more reliable. This finding is important because, if corpus similarity measures remain robust across different data sizes, we could predict the degree of confidence we should have for embeddings trained in truly low-resource contexts.

While this relationship is strong in general, there are some clear outliers: for example, in Arabic the correlation is only 0.567. This means that there is only a weak relationship between corpus homogeneity and the reliability of low-resource embeddings. Russian is another language with a relatively weak relationship, in this case 0.786. While Arabic is the only Afro-Asiatic language in this data set, the other Balto-Slavic language (Polish) has a much stronger relationship than Russian. One potential factor is that Arabic morphology, a unique root-and-pattern system, is not shared by any other language in this study. This raises a question for future work about whether related languages like Hebrew would evidence this same characteristic.

The experiment in this section has shown that most languages have a strong relationship between the homogeneity of the training corpus and the reliability of very low-resource embeddings (trained on only 1 million words). The implications of this relationship are important because it suggests that we can predict the reliability of low-resource embeddings.

\section{Experiment 4: Corpus Similarity}

The final experiment asks how much change we see in corpus similarity measures given the size of the data set. We have calculated corpus similarity by dividing a corpus into many chunks of 20k words and calculating pairwise similarity between these chunks. That means we are estimating the overall similarity by sampling a number of individual observations (200 as implemented here). Here we measure the stability of corpus similarity measures over corpus size, as shown for Arabic in Figure \ref{fig3}. In this figure, the y-axis represents corpus similarity and the x-axis represents the size of the corpus. Overall, the measures are quite stable.

\begin{table}[h]
\centering
\begin{tabular}{|l|c|c|c|c|}
\hline
\textbf{Language} & \textbf{Ftr.} & \textbf{\textsc{cc-tw}} & \textbf{\textsc{cc-wk}} & \textbf{\textsc{tw-wk}} \\
\hline
Arabic & C4 & 0.056 & 0.024 & 0.062 \\
German & C4 & 0.040 & 0.051 & 0.036 \\
Greek & W1 & 0.033 & 0.037 & 0.038 \\
Farsi & W1 & 0.041 & 0.023 & 0.061 \\
Finnish & C4 & 0.014 & 0.031 & 0.066 \\
French & W1 & 0.051 & 0.045 & 0.068 \\
Indonesian & C4 & 0.069 & 0.045 & 0.093 \\
Italian & W1 & 0.031 & 0.035 & 0.033 \\
Japanese & C2 & 0.025 & 0.028 & 0.038 \\
Korean & C4 & 0.046 & 0.063 & 0.058 \\
Dutch & W1 & 0.023 & 0.038 & 0.042 \\
Polish & W1 & 0.022 & 0.069 & 0.059 \\
Portuguese & C4 & 0.071 & 0.030 & 0.056 \\
Russian & C4 & 0.067 & 0.069 & 0.071 \\
Spanish & C4 & 0.019 & 0.040 & 0.055 \\
Swedish & C4 & 0.014 & 0.054 & 0.040 \\
Turkish & C4 & 0.045 & 0.028 & 0.104 \\
\hline
\textbf{Average} & \textbf{All} & \textbf{0.039} & \textbf{0.041} & \textbf{0.057} \\
\hline
  \end{tabular}
  \caption{$Max-Min$ of Similarity Across Sizes}
  \label{tab:6}
\end{table}

We explore the stability of corpus similarity measures in Table \ref{tab:6}. This table takes the average corpus similarity value across each amount of data (10 million words, 20 million words, and so on) and then finds the range between the maximum similarity and the minimum similarity for each condition: for example, the similarity between Arabic tweets and Arabic Wikipedia articles at different sample sizes. This table shows that there is only a small variation across sample size in each condition. We further test this using a one-sample t-test to see if each condition actually constitutes a single population. In all cases, there is no significant difference among the population of corpus similarity values by size, a confirmation of the visual trend in Figure \ref{fig3}. Thus, corpus similarity measures are robust regardless of the size of the corpus.

\section{Conclusions}

This paper has simulated low-resource settings in a cross-lingual and cross-register context in order to measure the similarity, stability, and reliability of embeddings. The basic idea has been to examine the ability of corpus similarity measures, applied to training data, to predict downstream differences in embeddings. The important background is that corpus similarity measures remain stable across corpus size, so that they can be applied in truly low-resource settings.

The first findings, in Sections 6 and 7, showed that (i) register-specific embeddings are significantly different and (ii) that embeddings become more stable within registers as the amount of training data increases. Both of these findings are expected. The important new contribution is the fact that the degree of difference in register-specific embeddings can in fact be predicted by differences in the training corpora themselves. Because low-resource languages have a reduced inventory of register-specific corpora, it is not possible to directly measure the impact of register on embeddings in such languages. Corpus similarity measures thus allow us to indirectly measure the impact of register in truly low-resource settings.

The second important finding, in Section 8, is that the stability of low-resource embeddings can be predicted given corpus homogeneity measures. In a truly low-resource setting, we would never be able to measure embedding reliability because of data limitations. We can, however, measure corpus homogeneity even with limited corpus sizes (c.f., Section 9). The combination of these two findings, then, means that it is possible to predict which low-resource embeddings are more reliable and which are less reliable. This constitutes a significant advance in validating low-resource language resources, providing a measure of confidence for embeddings trained from small corpora.

The caveat of the experiments in this paper, however, is that we have focused on \textit{simulated} low-resource settings rather than \textit{actual} low-resource settings. This is a necessary choice given the need to undertake a large number of comparisons within each language. Further, the 17 languages used here represent a range of language families and types of morphology, but we know that truly low-resource languages often belong to families that are not represented here.

A \href{https://www.github.com/jonathandunn/corpus_similarity}{\underline{Python package}} is made available for working with these corpus similarity measures and the full experimental results are available in the \href{https://www.dropbox.com/s/hsdzyatb4d2a3bm/LREC_2022.Supplementary_Material.zip?dl=0}{\underline{supplementary material}}.

\section{Bibliographical References}\label{reference}


\bibliographystyle{lrec2022-bib}
\bibliography{lrec2022-example}

\begin{thebibliography}{}

\bibitem[\protect\citename{Adelmann \bgroup et al.\egroup
  }2021]{adelmann-etal-2021-impact}
Adelmann, B., Menzel, W., and Zinsmeister, H.
\newblock (2021).
\newblock {The Impact of Word Embeddings on Neural Dependency Parsing}.
\newblock In {\em Proceedings of the 17th Conference on Natural Language
  Processing}, pages 1--13, D{\"{u}}sseldorf, Germany. KONVENS 2021 Organizers.

\bibitem[\protect\citename{Agi{\'{c}} and
  Vuli{\'{c}}}2019]{agic-vulic-2019-jw300}
Agi{\'{c}}, {\v{Z}}. and Vuli{\'{c}}, I.
\newblock (2019).
\newblock {JW300: A Wide-Coverage Parallel Corpus for Low-Resource Languages}.
\newblock In {\em Proceedings of the Annual Meeting of the Association for
  Computational Linguistics}, pages 3204--3210, Florence, Italy, jul.
  Association for Computational Linguistics.

\bibitem[\protect\citename{Ali}2011]{Ali2011}
Ali, A.
\newblock (2011).
\newblock {\em {Textual similarity}}.
\newblock Bachelors thesis, Technical University of Denmark.

\bibitem[\protect\citename{Antoniak and
  Mimno}2018]{antoniak-mimno-2018-evaluating}
Antoniak, M. and Mimno, D.
\newblock (2018).
\newblock {Evaluating the Stability of Embedding-based Word Similarities}.
\newblock {\em Transactions of the Association for Computational Linguistics},
  6:107--119.

\bibitem[\protect\citename{Biber and Conrad}2009]{Biber2009}
Biber, D. and Conrad, S.
\newblock (2009).
\newblock {\em {Register, genre, and style}}.
\newblock Cambridge University Press, Cambridge, UK.

\bibitem[\protect\citename{Biber \bgroup et al.\egroup }2020]{Biber2020}
Biber, D., Egbert, J., and Keller, D.
\newblock (2020).
\newblock {Reconceptualizing register in a continuous situational space}.
\newblock {\em Corpus Linguistics and Linguistic Theory}, 16(3):581--616.

\bibitem[\protect\citename{Biber}2012]{Biber2012}
Biber, D.
\newblock (2012).
\newblock {Register as a predictor of linguistic variation.}
\newblock {\em Corpus Linguistics and Linguistic Theory}, 8(1):9--37.

\bibitem[\protect\citename{Burdick \bgroup et al.\egroup
  }2021]{burdick-etal-2021-analyzing}
Burdick, L., Kummerfeld, J.~K., and Mihalcea, R.
\newblock (2021).
\newblock {Analyzing the Surprising Variability in Word Embedding Stability
  Across Languages}.
\newblock In {\em Proceedings of the 2021 Conference on Empirical Methods in
  Natural Language Processing}, pages 5891--5901, Online and Punta Cana,
  Dominican Republic, nov. Association for Computational Linguistics.

\bibitem[\protect\citename{Christodoulopoulos and
  Steedman}2015]{ChristodoulopoulosC.andSteedman2015}
Christodoulopoulos, C. and Steedman, M.
\newblock (2015).
\newblock {A massively parallel corpus: The Bible in 100 languages}.
\newblock {\em Language Resources and Evaluation}, 49(2):375--395.

\bibitem[\protect\citename{Dunn and Adams}2020]{dunn-adams-2020-geographically}
Dunn, J. and Adams, B.
\newblock (2020).
\newblock {Geographically-Balanced {G}igaword Corpora for 50 Language
  Varieties}.
\newblock In {\em Proceedings of the 12th Language Resources and Evaluation
  Conference}, pages 2528--2536, Marseille, France, may. European Language
  Resources Association.

\bibitem[\protect\citename{Dunn}2020]{Dunn2020}
Dunn, J.
\newblock (2020).
\newblock {Mapping languages: the Corpus of Global Language Use}.
\newblock {\em Language Resources and Evaluation}, 54:999--1018.

\bibitem[\protect\citename{Dunn}2021]{Dunn2021}
Dunn, J.
\newblock (2021).
\newblock {Representations of Language Varieties Are Reliable Given Corpus
  Similarity Measures}.
\newblock In {\em Proceedings of the Eighth Workshop on NLP for Similar
  Languages, Varieties and Dialects}, pages 28--38. Association for
  Computational Linguistics.

\bibitem[\protect\citename{Fothergill \bgroup et al.\egroup
  }2016]{fothergill-etal-2016-evaluating}
Fothergill, R., Cook, P., and Baldwin, T.
\newblock (2016).
\newblock {Evaluating a Topic Modelling Approach to Measuring Corpus
  Similarity}.
\newblock In {\em Proceedings of the Tenth International Conference on Language
  Resources and Evaluation ({LREC}'16)}, pages 273--279, Portoro{\v{z}},
  Slovenia, may. European Language Resources Association (ELRA).

\bibitem[\protect\citename{Hellrich \bgroup et al.\egroup }2019]{Hellrich2019}
Hellrich, J., Kampe, B., and Hahn, U.
\newblock (2019).
\newblock {The Influence of Down-Sampling Strategies on SVD Word Embedding
  Stability}.
\newblock In {\em Proceedings of the 3rd Workshop on Evaluating Vector Space
  Representations for NLP}, pages 18--26. Association for Computational
  Linguistics.

\bibitem[\protect\citename{Kilgarriff}1997]{kilgarriff-1997-using}
Kilgarriff, A.
\newblock (1997).
\newblock {Using Word Frequency Lists to Measure Corpus Homogeneity and
  Similarity between Corpora}.
\newblock In {\em Proceedings of the Fifth Workshop on Very Large Corpora},
  pages 231--245. Association for Computational Linguistics.

\bibitem[\protect\citename{Kilgarriff}2001]{Kilgarriff2001}
Kilgarriff, A.
\newblock (2001).
\newblock {Comparing Corpora}.
\newblock {\em International Journal of Corpus Linguistics}, 6(1):97--133.

\bibitem[\protect\citename{Kilgarriff}2009]{kilgarriff-2009}
Kilgarriff, A.
\newblock (2009).
\newblock {Simple maths for keywords}.
\newblock In {\em Proceedings of the Corpus Linguistics Conference}, Liverpool,
  UK. University of Liverpool.

\bibitem[\protect\citename{Leban \bgroup et al.\egroup
  }2016]{DBLP:conf/ecir/LebanFG16}
Leban, G., Fortuna, B., and Grobelnik, M.
\newblock (2016).
\newblock {Using News Articles for Real-time Cross-Lingual Event Detection and
  Filtering}.
\newblock In {\em Proceedings of the First International Workshop on Recent
  Trends in News Information Retrieval}, volume 1568, pages 33--38. CEUR
  Workshop Proceedings.

\bibitem[\protect\citename{Lu \bgroup et al.\egroup
  }2020]{lu-etal-2020-diverging}
Lu, J., Henchion, M., and {Mac Namee}, B.
\newblock (2020).
\newblock {Diverging Divergences: Examining Variants of {J}ensen {S}hannon
  Divergence for Corpus Comparison Tasks}.
\newblock In {\em Proceedings of the 12th Language Resources and Evaluation
  Conference}, pages 6740--6744, Marseille, France, may. European Language
  Resources Association.

\bibitem[\protect\citename{Miaschi and
  Dell'Orletta}2020]{miaschi-dellorletta-2020-contextual}
Miaschi, A. and Dell'Orletta, F.
\newblock (2020).
\newblock {Contextual and Non-Contextual Word Embeddings: an in-depth
  Linguistic Investigation}.
\newblock In {\em Proceedings of the 5th Workshop on Representation Learning
  for NLP}, pages 110--119, Online, jul. Association for Computational
  Linguistics.

\bibitem[\protect\citename{Mikolov \bgroup et al.\egroup
  }2018]{mikolov2018advances}
Mikolov, T., Grave, E., Bojanowski, P., Puhrsch, C., and Joulin, A.
\newblock (2018).
\newblock {Advances in Pre-Training Distributed Word Representations}.
\newblock In {\em Proceedings of the International Conference on Language
  Resources and Evaluation (LREC 2018)}, pages 52--55. European Language
  Resources Association.

\bibitem[\protect\citename{Nanayakkara and Ranathunga}2018]{Nanayakkara-2018}
Nanayakkara, P. and Ranathunga, S.
\newblock (2018).
\newblock {Clustering Sinhala News Articles Using Corpus-Based Similarity
  Measures}.
\newblock In {\em 2018 Moratuwa Engineering Research Conference (MERCon)},
  pages 437--442.

\bibitem[\protect\citename{Piperski}2017]{Piperski-2017}
Piperski, A.
\newblock (2017).
\newblock {Sum of Minimum Frequencies as a Measure of Corpus Similarity}.
\newblock In {\em Proceedings of the Corpus Linguistics Conference},
  Birmingham, UK.

\bibitem[\protect\citename{Tiedemann}2012]{Tiedemann2012}
Tiedemann, J.
\newblock (2012).
\newblock {Parallel Data, Tools and Interfaces in OPUS}.
\newblock In {\em Proceedings of the International Conference on Language
  Resources and Evaluation}, page 2214–2218. European Language Resources
  Association.

\bibitem[\protect\citename{Torres-Moreno \bgroup et al.\egroup
  }2014]{DBLP:journals/corr/Torres-MorenoSP17}
Torres-Moreno, J.-M., Sierra, G., and Peinl, P.
\newblock (2014).
\newblock {A German Corpus for Text Similarity Detection Tasks}.
\newblock {\em International Journal of Computational Linguistics and
  Applications}, 5:9--24.

\bibitem[\protect\citename{Wendlandt \bgroup et al.\egroup
  }2018]{wendlandt-etal-2018-factors}
Wendlandt, L., Kummerfeld, J.~K., and Mihalcea, R.
\newblock (2018).
\newblock {Factors Influencing the Surprising Instability of Word Embeddings}.
\newblock In {\em Proceedings of the 2018 Conference of the North {A}merican
  Chapter of the Association for Computational Linguistics: Human Language
  Technologies, Volume 1 (Long Papers)}, pages 2092--2102, New Orleans,
  Louisiana, jun. Association for Computational Linguistics.

\end{thebibliography}

\section{Language Resource References}
\label{lr:ref}

\hangindent=1cm Dunn, J.; Li, H.; \& Sastre, D. (2022). \textit{Corpus Similarity: A Python package}. \href{https://github.com/jonathandunn/corpus_similarity}{https://github.com/jonathandunn/corpus\_similarity}.


\end{document}